\documentclass[runningheads]{llncs}

\usepackage{times}
\usepackage{epsfig}
\usepackage{graphicx}
\usepackage{amsmath}
\usepackage{amssymb}
\usepackage{mathrsfs}

\usepackage[utf8]{inputenc}

\usepackage[normalem]{ ulem }
\usepackage{soul}

\newcommand{\Rd}{\mathbb{R}^d}
\usepackage{algorithm}
\usepackage{algorithmic}

\usepackage{subfig}
\usepackage{xcolor}

\usepackage{authblk}

\newcommand{\dps}{\displaystyle}
\newcommand{\om}{\omega}
\newcommand{\ebe}{\,|\,}

\newcommand{\R}{\mathbb{R}}

\usepackage[breaklinks=true,bookmarks=false]{hyperref}

        \usepackage{dutchcal}

\newcommand{\x}{\vec{\mathbf{x}} } 
\newcommand{\y}{\vec{\mathbf{y}}} 
\newcommand{\T}{\vec{\mathbf{T}} } 

\newcommand{\Src}{S} 
\newcommand{\Tgt}{T} 
\newcommand{\Srcn}{{S_n}}

\newcommand{\omg}{\omega}
\newcommand{\Det}[1]{\left| #1 \right|}

\newcommand{\USd}{\mathbb{S}^{d-1}}
\newcommand{\RxSd}{\Rd \times \USd}

\newcommand\restr[2]{\ensuremath{{#1}_{|#2}}}

\newtheorem{Proposition}{Proposition} 
\newtheorem{Definition}{Definition} 

\title{Partial Matching in the Space of Varifolds}

\author{****\inst{1}}
\institute{ \email{****@**.***}}

\author{Pierre-Louis Antonsanti\inst{1,2} \and
Joan Glaunès\inst{2} \and
Thomas Benseghir\inst{1} \and 
Vincent Jugnon\inst{1} \and
Irène Kaltenmark\inst{2}}

\authorrunning{P. Antonsanti et al.}

\institute{GE Healthcare, Buc 78530, France \\ \email{\{pierrelouis.antonsanti,thomas.benseghir,vincent.jugnon\}@ge.com} \and
MAP5, Universite de Paris, Paris 75006, France \email{alexis.glaunes@parisdescartes.fr}}

\begin{document}

\maketitle

%
\begin{abstract}
    In computer vision and medical imaging, the problem of matching structures finds numerous applications from automatic annotation to data reconstruction. 
    The data however, while corresponding to the same anatomy, are often very different in topology or shape and might only partially match each other.
    We introduce a new asymmetric data dissimilarity term for various geometric shapes like sets of curves or surfaces. 
    This term is based on the Varifold shape representation and assesses the embedding of a shape into another one without relying on correspondences between points.
    It is designed as data attachment for the Large Deformation Diffeomorphic Metric Mapping (LDDMM) framework, allowing to compute meaningful deformation of one shape onto a subset of the other.
    Registrations are illustrated on sets of synthetic 3D curves, real vascular trees and livers' surfaces from two different modalities: Computed Tomography (CT) and Cone Beam Computed Tomography (CBCT).
    All experiments show that this data dissimilarity term leads to coherent partial matching despite the topological differences.
\end{abstract}

\section{Introduction}

Finding shape correspondences is a standard problem in computer vision that has numerous applications such as pattern recognition (\cite{Bronstein2006}, \cite{Bronstein2009}, \cite{Kaick2013}), annotation (\cite{Benseghir2013}, \cite{Feragen2015}) and reconstruction (\cite{Halimi2020}).
In particular, in the field of medical imaging, matching an atlas and a patient's anatomy (\cite{Feragen2015}),
or comparing exams of the same patient acquired with different imaging techniques (\cite{Zhen2012}, \cite{Bashiri2018}), provide critical information to physicians for both planning and decision making. 

In medical imaging, this problem has been tackled by numerous authors (\cite{Sotiras2013}) by registering directly the images, most of the time assuming that both images contains the entire object of interest.
But in practice, it often happens that only part of an object is visible in one of the two modalities: 
in CBCT for instance, the imaged organs can be larger than the field of view, when the whole organ can be acquired in CT. 
In order to make the best of the two modalities, one needs to find a partial matching between them. 
This work is focused on sparse segmented structures registration where only part of these structures can be matched. 

\subsubsection{Previous works.}
The problem of matching shapes has been widely addressed in the literature in the past decades (\cite{Kaick2011}). In the specific case of partial matching, one can find two main approaches to such problem: either by finding correspondences, sparse or dense, between structures from descriptors that are invariant to different transformations (\cite{Aiger2008}, \cite{Rodola2017}, \cite{Halimi2019}),
or by looking for a deformation aligning the shapes with respect to a given metric (\cite{Aiger2008}, \cite{Benseghir2013}, \cite{Feragen2015}, \cite{Halimi2020}).

The early works on partial shape correspondence as reviewed in \cite{Kaick2011} rely on {\bf correspondences} between points computed from geometric descriptors extracted from an isotropic local region around the selected points. 
The method is refined in \cite{Kaick2013} by selecting pairs of points to better fit the local geometry using bilateral map.
The features extracted can also be invariant to different transformation, as in \cite{Rodola2013} where the descriptors extraced are scale invariant. 
Such sparse correspondences are naturally adapted to partial matching, yet they cannot take the whole shapes into account in the matching process. 

Using a different approach, functional maps were introduced in \cite{Ovsjanikov2013} allowing dense correspondences between shapes by transferring the problem to linear functions between spaces of functions defined over the shapes. 
In \cite{Rodola2017} the non-rigid partial shape correspondence is based on the Laplace-Beltrami eigenfunctions used as prior in the spectral representation of the shapes. 
Recently in \cite{Halimi2019}, such functional map models were adapted in a deep unsupervised framework by finding correspondences minimizing the distortion between shapes. Such methods are yet limited to surface correspondences. 

 The second kind of approaches relies on the {\bf deformations} that can be generated to align the shapes with each other. 
It usually involves minimizing a function called data attachment that quantifies the alignment error between the shapes. 
A deformation cost is sometimes added to regularize these deformations. 
The sparse correspondences being naturally suited to partial matching, they are notably used in Iterative Closest Point (ICP) methods and their derivatives to guide a registration of one shape onto the other. 
In \cite{Aiger2008} a regularized version of the ICP selects the sets of four co-planar points in the points cloud. 
In \cite{Benseghir2013} the ICP is adapted to the specific case of vascular trees and compute curves' correspondences through an Iterative Closest Curve method. 
Working on trees of 3D curves as well, \cite{Feragen2015} hierarchically selects the overall curves correspondences minimizing the tree space geodesic distance between the trees. 
This latter method although specific to tree-structures, allows topological changes in the deformation. 

On the other hand some authors compute the deformation guided by a dense data attachment term. In \cite{Bronstein2006} isometry-invariant minimum distortion deformations are applied to 2D Riemannian manifolds thanks to a multiscale framework using partial matching derived from Gromov's theory. This work is extended in \cite{Bronstein2009}  by finding the optimal trade-off between partial shape matching and similarity between these shapes embedded in the same metric space.
Recently in \cite{Halimi2020} a partial correspondence is performed through a non-rigid alignment of one shape and its partial scan seen as points clouds embedded in the same representation space. 
The non-rigid alignment is done with a Siamese architecture network. 
This approach seems promising and is part of a completion framework, however it requires a huge amount of data to train the network. 

Interestingly, the regularization cost can be seen as a distance between shapes itself (as in \cite{Feragen2015}) by quantifying the deformation amount necessary to register one shape onto the other. This provides a complementary tool to the metrics used to quantify the shapes dissimilarities. 

A well established and versatile framework to compute meaningful deformations is the Large Deformation Diffeomorphic Metric Mapping. 
It allows to see the difference between shapes through the optimal deformations to register a source shape $\Src$ onto a target shape $\Tgt$. 
However, the data attachment metrics proposed so far aim to compare the source and target shapes in their entirety (\cite{Charon2020}), or look for explicit correspondences between subparts of these shapes (\cite{Feydy2017}).
In \cite{Kaltenmark2018} the growth model presented introduces a first notion of partial matching incorporated to the LDDMM framework, yet no explicit partial dissimilarity term was proposed.

We propose a new asymmetric dissimilarity term adapted as data attachment in the LDDMM framework derived from the Varifold shape representation \cite{Charon2013} to quantify the partial shape matching. 
This term can be used in both continuous and discrete settings, and applied to many discrete shape representations without requiring any point correspondences.
When combined with LDDMM, this approach allows to build meaningful deformations while being adapted to the partial matching between shapes.


\section{Partial Matching}

We are interested in the problem of finding an optimal deformation to register a source shape $\Src$ onto an unknown subset of a target shape $\Tgt$, where $\Src, \Tgt$ are assumed to be finite unions of $m$-dimensional submanifolds of $E=\Rd$, with either $m=1$ (curves) or $m=d-1$ (hypersurfaces). 

\subsection{The varifold framework for shape matching}

In this section we will quickly review the oriented varifold framework introduced in \cite{Kaltenmark2017}, of which the varifold framework (\cite{Charon2013}) is a particular case.

Let $E=\R^d$, with $d\geq2$ the ambient space.
Shapes $\Src$ and $\Tgt$ are assumed to be compact $m$-rectifiable subsets of $\Rd$. In particular at almost every point $x\in \Src$ (resp. $y\in \Tgt$) is defined a unit tangent - for curves - or normal - for hypersurfaces - vector $\tau_x\Src\in \USd$ (resp. $\tau_y\Tgt$). 

Let $W$ be a Reproducing Kernel Hilbert Space (RKHS) of functions defined over $\RxSd$, continuously embedded in $C_0(\RxSd)$. Its dual space $W'$ is a space of varifolds. The following proposition gives a practical way to define such a space:

\begin{Proposition}[\cite{Charon2013}, Lemma 4.1]
\label{tensor_product}
Assume that we are given a positive-definite real kernel $k_e$ on the space $\Rd$ such that $k_e$ is continuous, bounded and for all $x\in \Rd$, the function $k_e(x,.)$ vanishes at infinity. Assume that a second positive-definite real kernel $k_t$ is defined on the manifold $\USd$ and is also continuous. Then the RKHS W associated to the positive-definite kernel $k_e\otimes k_t$ is continuously embedded into the space $C_0(\RxSd)$.
\end{Proposition}

In the following we assume that the reproducing kernel of $W$ is of the form $k_e\otimes k_t$, with the assumptions of proposition \ref{tensor_product}. We also assume that $k_e$ and $k_t$ are non negative functions. 
In practice, we will use for $k_e:{(\Rd)}^2\rightarrow\R$ a gaussian kernel $k_e(x,y)=e^{-\|x-y\|^2/\sigma_W^2}$, where $\sigma_W$ is a scale parameter, and for $k_t:{(\USd)}^2\rightarrow\R$ the kernel $k_t(u,v) = e^{\langle\,u,v\rangle_{\Rd}}$

We associate with shape $\Src$ the canonical function $\omega_\Src\in W$ defined for all $y\in \Rd$ and $\tau \in \USd$ as follows: 
$$\omega_\Src(y,\tau)=\int_\Src k_e(y,x)k_t(\tau,\tau_x\Src)dx \,.$$
This function corresponds to the unique representer of the varifold $\mu_\Src\in W'$ via the Riesz representation theorem. Similarly, we define the canonical function $\omega_\Tgt$ associated with shape $\Tgt$. Via this representation of shapes, one may express the scalar product between the varifolds $\mu_\Src, \mu_\Tgt$, or equivalently between the canonical functions 
$\omega_\Src, \omega_\Tgt$ as follows:
\begin{eqnarray*}
\langle\,\mu_\Src,\mu_\Tgt\rangle_{W'}
=\langle\,\omega_\Src,\omega_\Tgt\rangle_{W}
=\int_\Src\int_\Tgt k_e(x,y)k_t(\tau_x\Src,\tau_{y}\Tgt)dx\,dy\,.
\end{eqnarray*}
Finally, the shape matching distance as defined in \cite{Charon2013} is the following:
\begin{eqnarray*}
d_{W'}(\Src,\Tgt)^2 &=& \Vert \omega_{\Src} - \omega_{\Tgt} \Vert_{W}^2 = \Vert \mu_{\Src} - \mu_{\Tgt} \Vert_{W'}^2 \\&=& \langle\,\mu_{\Src},\mu_{\Src}\rangle_{W'}-2\langle\,\mu_{\Src},\mu_{\Tgt}\rangle_{W'}+\langle\,\mu_{\Tgt},\mu_{\Tgt}\rangle_{W'}\,.
\end{eqnarray*}
In order to adapt this distance for partial matching, a first and intuitive way could be to use half of the expression as follows:
$$\Delta(\Src,\Tgt) = \langle\,\mu_{\Src},\mu_{\Tgt} - \mu_{\Src}\rangle_{W'}^2 = \langle\,\mu_{\Src},\mu_{\Src}-\mu_{\Tgt}  \rangle_{W'}^2 .$$
The intuition behind this definition is that if $\Src$ is a subset of $\Tgt$, then $\mu_\Tgt-\mu_\Src$ is the varifold corresponding to $\Tgt\setminus\Src$, which is disjoint from shape $\Src$ and thus roughly orthogonal to it from the varifold metric viewpoint.

Yet, if $S = S_1 \sqcup S_2$, a mismatch of $S_1$ into $T$, characterized by $\langle\,\mu_{\Src_2}, \mu_{\Src}-\mu_{\Tgt}\rangle_{W'} > 0$, can be compensated by an overrated characterization of the inclusion $S_2 \subset T$ with $\langle\,\mu_{\Src_2}, \mu_{\Src} -\mu_{\Tgt}\rangle_{W'} < 0$, which happens if the mass of $T$ around ${\Src_2}$ is larger than the mass of ${\Src_2}$. Hence, we introduce in this paper a localized characterization of the inclusion.

\subsection{Definition of the partial matching dissimilarity}

To simplify the notation, we denote for $x,x' \in S$, $\x = (x,\tau_x S)$, $\om_S(\x)=\omega_S(x,\tau_x\Src)$ and $k(\x,\x') = k_e(x,x')k_t(\tau_x\Src,\tau_{x'}\Src)$. 

\begin{Definition}\label{def:partial_dissimilarity}
Let $g : \mathbb{R} \mapsto \mathbb{R}$ defined as $g(s) = (\max(0,s))^2$.
We define the partial matching dissimilarity as follows: 
\begin{eqnarray*}
\Delta(\Src,\Tgt) &=&
 \int_{\Src}g\left(\om_S(\x) - \om_\Tgt(\x)\right)dx
\\&=&\int_{\Src}g\left(\int_\Src k(\x,\x')dx'-\int_\Tgt k(\x,\y)dy\right)dx\,.
\end{eqnarray*}

\end{Definition}
With $g(s) = s$, we would retrieve $\langle\,\mu_{\Src},\mu_{\Src}-\mu_{\Tgt}  \rangle_{W'}$. The threshold $\max(0,\cdot)$ prevents the compensation of a local mismatch by an overrated match in another area. 

\begin{Proposition}\label{prop:inclusion1}
If $\Src \subset \Tgt$, $\Delta(\Src,\Tgt)=0$.
\end{Proposition}

Since $k_e$ and $k_t$ are assumed to be non negative functions, we have
\begin{eqnarray}
\Delta(\Src,\Tgt) &=&\int_{\Src}g\left(\int_\Src k(\x,\y)dy-\int_\Tgt k(\x,\y)dy\right)dx  \nonumber\\
&=&\int_{\Src}g\left(-\int_{{\Tgt \setminus \Src}} k(\x,\y)dy\right)dx
\label{eq:Tprime}
=0\,.\quad \qed
\end{eqnarray}

The next proposition highlights the local nature of the dissimilarity function $\Delta$.
\begin{Proposition}\label{pro:croissance}
If $\Src' \subset \Src$, then $\Delta(\Src',\Tgt) \leq \Delta(\Src,\Tgt)$. In particular, if \\ ${\Delta(\Src,\Tgt)=0}$ then for any subset $\Src'$ of $\Src$, $\Delta(\Src',\Tgt)=0$.
\end{Proposition}
Since $k \geq 0$, we have for any $\Src' \subset \Src$ and any $\y \in \RxSd$ 
$$\om_{\Src'}(\y)=\int_{\Src'}  k(\x,\y) dx \leq \int_{\Src} k(\x,\y)dx = \om_{\Src}(\y)\,.$$
Hence, since $g$ is an increasing function $${g\left(\omega_{\Src'}(\y) - \omega_\Tgt(\y)\right) \leq g\left(\omega_\Src(\y) - \omega_\Tgt(\y)\right)}$$
and thus $\dps \Delta(\Src',\Tgt)\leq \int_{\Src}g\left(\omega_\Src(\x) - \omega_\Tgt(\x)\right)dx =\Delta(\Src,\Tgt)$. $\quad \qed$

In the next proposition, we show that $\Delta$ does not satisfies the property\\ $\Delta(\Src,\Tgt) = 0 \Rightarrow \Src \subseteq \Tgt$. 
\begin{Proposition}\label{prop:inclusion}
Consider the two following shapes. The source is a segment \\ $S_\epsilon= \{ (s,\epsilon) \ebe s \in [-\alpha,\alpha]\}$ slightly shifted by a step $\epsilon>0$ above a larger target $T=\{ (t,0) \ebe t \in [-\beta,\beta]\}$, with $0<\alpha < \beta$. Since the tangent vectors are almost all equal, we can ignore $k_t$ and consider a kernel $k_e$ defined by a decreasing function $\rho : \R^+ \to \R^+$ as follows $k_e(x,x') = \rho(|x-x'|^2)$. Then for any such $\rho$, there exists $(\epsilon,\alpha,\beta)$ such that $\Delta(S_\epsilon^\alpha,T^\beta) = 0$ and $S_\epsilon^\alpha \cap T^\beta = \emptyset$.
\end{Proposition}
We need to show that for any $x_0 \in S$, $\om_S(x_0) \leq \om_T(x_0)$ where \\$\om_S(x_0) = \int_S k_e(x_0,x)dx$ and $\om_T(x_0) = \int_T k_e(x_0,x')dx'$. Denote $x = (s,\epsilon) \in S$, $x_0 = (s_0,\epsilon) \in S$ and $x' = (t,0) \in T$ then $\|x-x_0\|^2 = \|(s,\epsilon)-(s_0,\epsilon)\|^2 = (s-s_0)^2$ and $\|x'-x_0\|^2 = \|(t,\epsilon)-(s_0,\epsilon)\|^2 = (t-s_0)^2 + \epsilon^2$. We then obtain
$$\om_S(x_0) = \int_{-\alpha}^\alpha \rho((s-s_0)^2) ds, \qquad \qquad \om_T(x_0) = \int_{-\beta}^\beta \rho((s-s_0)^2 + \epsilon^2) ds \,.$$
Denote these integrals $I_\alpha(s_0) = \int_{-\alpha}^\alpha \rho((s-s_0)^2) ds$ and $I_\beta(s_0,\epsilon) =\int_{-\beta}^\beta \rho((s-s_0)^2 + \epsilon^2) ds$. The integrands are symmetric with respect to $s=s_0$ and since $\rho$ is decreasing, we have the following inequalities:
\begin{align}
\mbox{for any } s_0 \in [-\alpha,\alpha], \qquad &I_\alpha(\alpha) \leq I_\alpha(s_0) \leq I_\alpha(0),\label{eq:prop_ineq1}\\
\mbox{for any } s_0 \in [-\alpha,\alpha], \mbox{for any } \epsilon>0, \qquad &I_\beta(\alpha,\epsilon) \leq I_\beta(s_0,\epsilon) \leq I_\beta(0,\epsilon).\label{eq:prop_ineq2}
\end{align}
Let us now show that there exist $(\epsilon,\alpha,\beta)$ such that $I_\alpha(0) \leq I_\beta(\alpha,\epsilon) $ that is\\
$\int_{-\alpha}^\alpha \rho(s^2) ds \leq \int_{-\beta}^\beta \rho((s-\alpha)^2 + \epsilon^2) ds$.\\
For $\alpha$ small enough and $\beta$ large enough, $\int_{-\beta}^\beta \rho((s-\alpha)^2 + \epsilon^2) ds \geq \int_{-2\alpha}^{2\alpha} \rho(s^2 + \epsilon^2) ds$. This last integral tends to $\int_{-2\alpha}^{2\alpha} \rho(s^2) ds$ when $\epsilon$ tends to $0$ and this limit is strictly larger than $I_\alpha(0)$ (with $\alpha$ small enough, $\rho(\alpha)>0$). Thus, for $\epsilon$ small enough, we have $I_\alpha(0) < I_\beta(\alpha,\epsilon) $.\\
Thanks to eq. \eqref{eq:prop_ineq1} and \eqref{eq:prop_ineq2}, we deduce that for any $x_0 \in S$, $\om_S(x_0) \leq \om_T(x_0)$. $\quad \qed$\\

This example shows that if the mass of the target is larger than the mass of the source then this excess of mass can compensate the lack of alignment between the shapes. For this reason, we introduce a normalized dissimilarity term.

\subsection{Normalized partial matching dissimilarity}

Assume that $x_0 \in S$ and $y_0 \in T$ are two close points. If around these points, the mass of $T$ is twice the mass of $S$, i.e. $\om_S(\x_0) \approx \frac{1}{2} \om_T(\y_0)$, then the local embedding of $S$ in $T$ is characterized by  $\om_S(\x_0) \leq \frac{1}{2} \om_T(\x_0)$ and more generally by $\om_S(\x_0) \leq \frac{\om_S(\x_0)}{\om_T(\y_0)} \om_T(\x_0)\label{test}$. Conversely, if the mass of $S$ is twice the mass of $T$, then we consider that locally $S \varsubsetneq T$ (e.g. two branches of a tree should not match the same branch of a target). Hence, the criterion of Definition \ref{def:partial_dissimilarity} should be preserved : $\om_S(\x_0) \leq \om_T(\x_0)$ is not satisfied. These observations lead to a new dissimilarity term that encompasses these two cases.

\begin{Definition}\label{def:partial_normalized_dissimilarity}
Using the same threshold function $g$ as in Definition \ref{def:partial_dissimilarity}, we define the partial matching \textbf{normalized} dissimilarity as follows: 

\begin{eqnarray*}
\underline{\Delta}(\Src,\Tgt) &=& 
 \int_{\Src}g\left(\om_S(\x)- \int_\Tgt     {\min}_\epsilon \left( 1, \dfrac{\om_{\Src}(\x)}{\om_\Tgt(\y)} \right) k(\x,\y) dy \right) dx
\end{eqnarray*}
where $\min_\epsilon(1,s)=\frac{s+1-\sqrt{\epsilon+(s-1)^2}}{2}$ with $\epsilon>0$ small, is used as a smooth approximation of the $\min(1,\cdot)$ function.
\end{Definition}

\subsection{Use in the LDDMM setting}\label{sec:LDDMM}
The framework we propose is sufficiently flexible to be embedded in a variety of inexact registration methods; in this paper, we focus on the LDDMM model described in \cite{Beg2005}. In this model, diffeomorphisms are constructed as flows of time-dependent square integrable velocity fields $t \in [0,1] \mapsto v_t$, each $v_t$ belonging to a predefined Hilbert space $V$ of smooth vector fields.
In the following we will denote $\phi^v_1$ the diffeomorphism of $\R^d$, solution at $t=1$ of the flow equation $\partial_t \phi^v_t = v_t \circ \phi^v_t$
with initial condition $\phi^v_0 = id$. 

\begin{Proposition}
Let $\lambda>0$ be a fixed parameter. The partial matching problem, which consists in minimizing over $L_V^2$ the function :
$$ J(v) = \lambda \int_0^1 \Vert v_t \Vert_V^2dt + \underline{\Delta}({\phi_1^v(\Src)},\Tgt)\,,$$
has a solution.
\end{Proposition}
From \cite{Glaunes2005}, theorem 7, the proof boils down to showing that the mapping 
$ v\mapsto A(v) = \underline{\Delta}({\phi_1^v(S)},T), $ is weakly continuous on $L_V^2$. 
Let $(v_n)$ be a sequence in $L^2_V$, weakly converging to some $v\in L^2_V$.
We need to show that $\underline{\Delta}(\phi_1^{v_n}(\Src) , \Tgt) \longrightarrow \underline{\Delta}(\phi_1^{v}(\Src) , \Tgt)$.\\
To simplify we denote $ \Src_n=\phi_1^{v_n}(\Src)$, $\Src_*=\phi_1^{v}(\Src)$ and for any $\x \in \RxSd$, $\dps f_n(\x) = \omega_\Srcn(\x) - \int_T {\min}_\epsilon \left( 1, \dfrac{\om_{\Srcn}(\x)}{\om_\Tgt(\y)} \right) k(\x,\y) dy$ and $f_*(\x)$ likewise for $\Src_*$.\\
We then have
\begin{equation}\label{eq:continuity}
    \begin{aligned} 
        \underline{\Delta}(\Srcn,\Tgt) - \underline{\Delta}({\Src_*},\Tgt) & = \mu_{\Src_n}(g \circ f_n) - \mu_{\Src_*} (g \circ f_*)\,.
    \end{aligned}
\end{equation}
The area formula
\begin{align*}
\mu_{\phi(S)} (\omg)= \int_S \omg(\phi(x),d_x\phi(\tau_x S)) \Det{\restr{d_x\phi}{\tau_x S}}dx,
\end{align*}
leads to
\begin{align*}
& \Big|\mu_{\Src_n}(g \circ f_n) - \mu_{\Src_*} (g \circ f_*)\Big| \leq \int_S  \Big|g\circ f_n(\phi^n(x),d_x\phi^n(\tau_x S)) \cdot \Det{\restr{d_x\phi^n }{\tau_x S}}  \\
& - g\circ f_*(\phi(x),d_x\phi(\tau_x S)) \cdot \Det{\restr{d_x\phi }{\tau_x S}}  \Big| dx\\
\leq \int_S & \Big|g\circ f_n(\phi^n(x),d_x\phi^n(\tau_x S)) \cdot \Det{\restr{d_x\phi^n }{\tau_x S}} - g\circ f_n(\phi(x),d_x\phi(\tau_x S)) \cdot \Det{\restr{d_x\phi }{\tau_x S}}  \\
& +g\circ f_n(\phi(x),d_x\phi(\tau_x S)) \cdot \Det{\restr{d_x\phi }{\tau_x S}} - g\circ f_*(\phi(x),d_x\phi(\tau_x S)) \cdot \Det{\restr{d_x\phi }{\tau_x S}}  \Big| dx\\
\leq \int_S & |g\circ f_n|_\infty  \cdot  \Big| \Det{\restr{d_x\phi^n }{\tau_x S}} - \Det{\restr{d_x\phi }{\tau_x S}} \Big| + \Det{d_x\phi}_\infty  |g\circ f_n - g\circ f_*|_\infty \, dx \,.
\end{align*}
Since $d_x\phi^n$ converge to $d_x\phi$, uniformly on $x\in \Src$ (\cite{Glaunes2005}), we only need to show that $ |g\circ f_n - g\circ f_*|_\infty  \to 0$. We first show that $ |f_n - f_*|_\infty  \to 0$. For any $\x \in \RxSd$
\begin{align}
f_n(\x) - f_*(\x) = &\om_{\Src_n}(\x) - \om_{\Src_*}(\x) \\
&+ \int_T k(\x,\y) \left[ {\min}_\epsilon \left( 1, \dfrac{\om_{\Srcn}(\x)}{\om_\Tgt(\y)} \right) - {\min}_\epsilon \left( 1, \dfrac{\om_{\Src_*}(\x)}{\om_\Tgt(\y)} \right) \right] dy \,.
\end{align}
Since $W$ is continuously embedded in $C_0^2(\RxSd)$, there exists $c_W$ such that for any $n$, $|\om_{\Src_n} - \om_{\Src_*}|_\infty \leq c_W |\om_{\Src_n} - \om_{\Src_*}|_W$. Moreover, since $v_n$ weakly converges to $v$, Corollary 1 from \cite{Charlier2017} ensures that $|\om_{\Src_n} - \om_{\Src_*}|_W \to 0$.\\
Regarding the integral, since $\R \ni s \mapsto \min_\epsilon(1,s)$ is Lipschitz, there exists $c_\epsilon>0$ such that
\begin{align}
&\left|\int_T k(\x,\y) \left[ {\min}_\epsilon \left( 1, \dfrac{\om_{\Srcn}(\x)}{\om_\Tgt(\y)} \right) - {\min}_\epsilon \left( 1, \dfrac{\om_{\Src_*}(\x)}{\om_\Tgt(\y)} \right) \right] dy \right| \\
&\leq \int_T \dfrac{k(\x,\y)}{|\om_T(\y)|} c_\epsilon |\om_{\Src_n}(\x) - \om_{\Src_*}(\x)| dy  \leq   c_\epsilon c_W |\om_{\Src_n} - \om_{\Src_*}|_W \int_T \dfrac{k(\x,\y)}{\om_T(\y)} dy
\end{align}
Since $T$ is compact and $\om_T$ is continuous and strictly positive on $\T = \{ (y, \tau_y T) \ebe y \in T \}$, we have $c_T = \inf_{\T} \om_T(\y) > 0$ so that $\dps \int_T \frac{k(\x,\y)}{\om_T(\y)} dy \leq \frac{\om_T(\x)}{c_T} \leq \frac{c_W |\om_T|_W}{c_T} < +\infty$. This shows that $ |f_n - f_*|_\infty  \to 0$. Now, since $f_*$ is bounded, there exists $M>0$ such that for any $n$, $|f_*|_\infty+|f_n|_\infty \leq M$ and since $g$ is locally Lipschitz, we deduce that $ |g\circ f_n - g\circ f_*|_\infty  \to 0\,.$ $\quad \qed$

\subsubsection{Discrete formulation.}
The discrete versions of the partial matching dissimilarities can be derived very straightforwardly, following the same discrete setting described in \cite{Charon2020}
for varifold matching. We omit its complete description here. The LDDMM registration procedure is numerically solved via a geodesic shooting algorithm \cite{Miller06}, optimizing on a set of initial momentum vectors located at the discretization points of the source shape.

\section{Experiments}
In order to evaluate the proposed dissimilarity terms, we conducted three experiments on two different types of data : two on sets of 3D-curves (one synthetic and one real) and one on surfaces. 
In all the following experiments, we initialize the registration by aligning objects barycenters since no prior positioning is known in our applications.
To model non-rigid deformations, we define the reproducing kernel $K_V$ of $V$ to be a sum of Gaussian kernels $K_V(x,y) = \sum_s \exp\left(-\Vert x-y\Vert^{2}\;/\;(\sigma_0/s)^{2}\right)$, where $s \in [1,4,8,16]$ and $\sigma_0$ is about half the size of the shapes bounding boxes.
For each set of experiments we use the same hyperparameters ($\sigma_0, \sigma_W, \lambda$) to compare the influence of the data attachment terms.  
Our Python implementation makes use of the libraries PyTorch \cite{pytorch} and \href{https://www.kernel-operations.io/keops/index.html}{KeOps} \cite{Keops2020}, to benefit from automatic differentiation and GPU acceleration of kernel convolutions.
\begin{figure*}[!ht]
\centering
   \subfloat[\label{fig:Init}]{%
      \includegraphics[trim={7cm 4.5cm 5cm 3cm}, clip, width=0.2\textwidth]{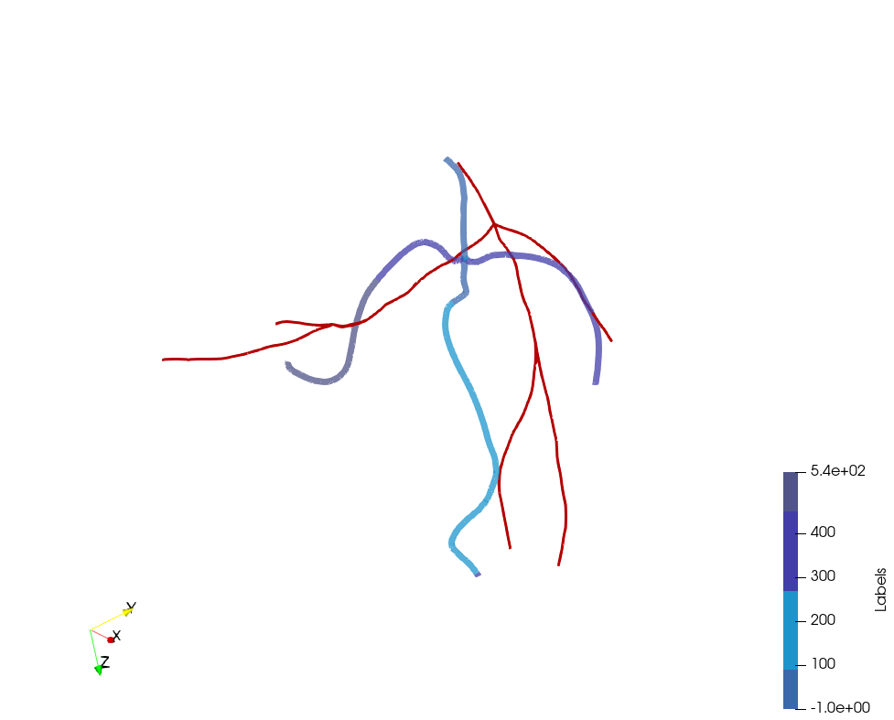}}
\hspace{.5cm}
\centering
   \subfloat[\label{fig:VarifoldSynthetic}]{%
      \includegraphics[trim={7cm 4.5cm 5cm 3cm}, clip, width=0.2\textwidth]{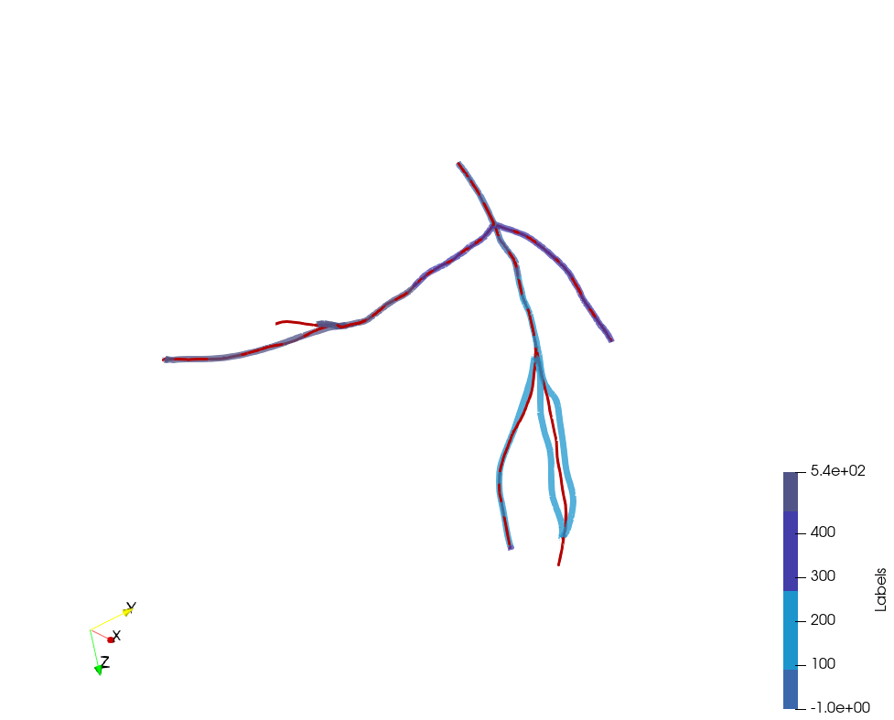}}
\hspace{.5cm}
\centering
   \subfloat[\label{fig:PartialVarifoldLocalSynthetic}]{%
      \includegraphics[trim={7cm 4.5cm 5cm 3cm}, clip, width=0.2\textwidth]{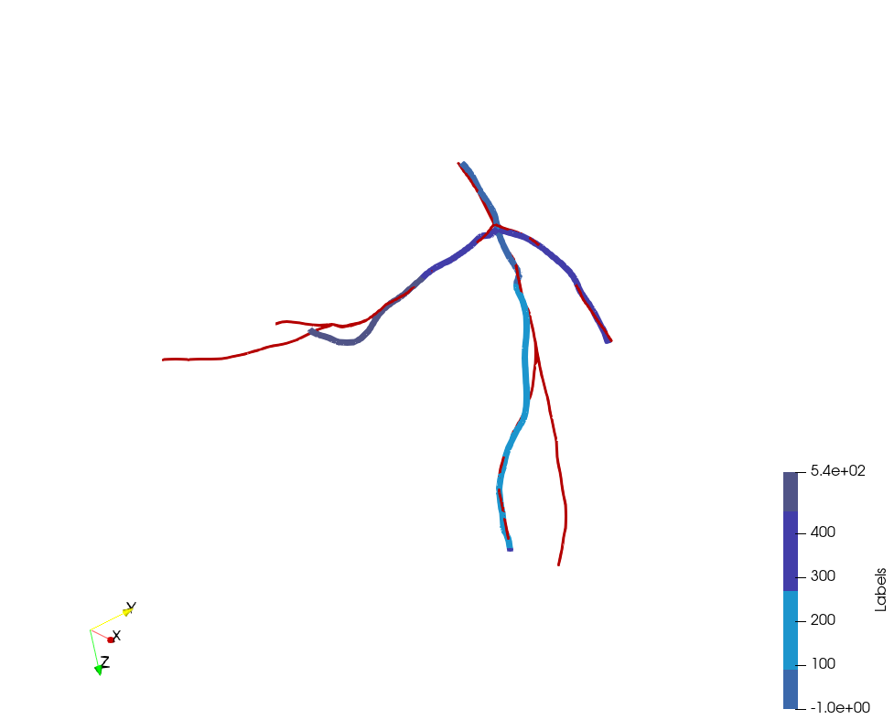}}
\hspace{.5cm}
\centering
   \subfloat[\label{fig:PartialVarifoldLocalNormalizedSynthetic}]{%
      \includegraphics[trim={7cm 4.5cm 5cm 3cm}, clip, width=0.2\textwidth]{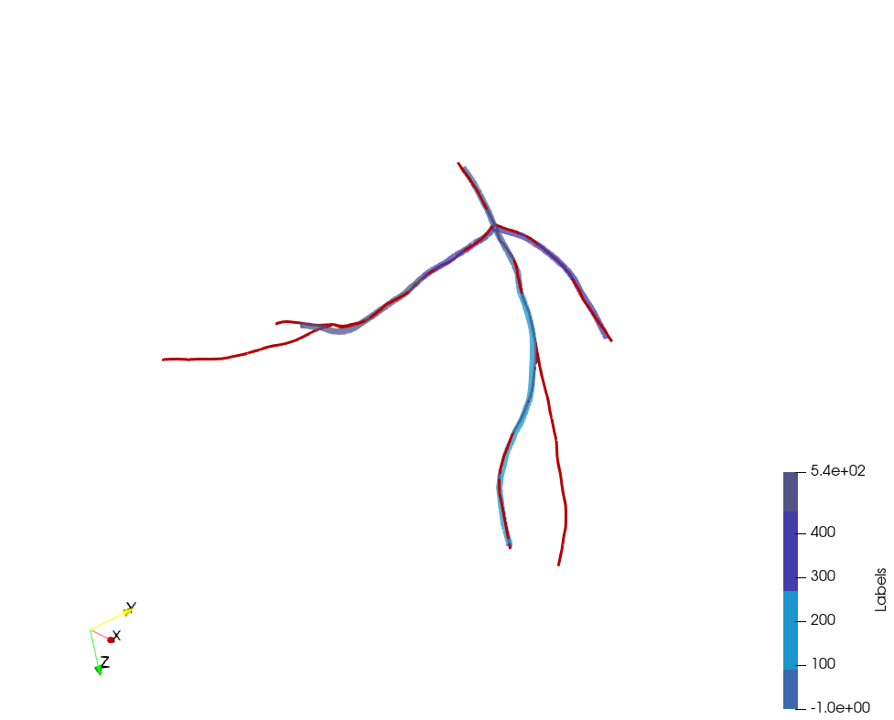}}\\
\caption{\label{fig:synthetic} Diffeomorphic registrations of a trimmed tree (blue) onto a richer one (red). (a) Initial Positions; (b) Varifold registration; (c) Partial matching registration
(Eq. \ref{def:partial_dissimilarity}); (d) Normalized partial matching registration (Eq. \ref{def:partial_normalized_dissimilarity}).}
\end{figure*}

\noindent\textbf{Synthetic experiment.}
A first experiment was conducted to validate our approach on synthetic trees of 3D-curves. 
The target is composed of six 3D-curves (red tree in Fig.\ref{fig:synthetic}.(a)), while the source (blue colored tree in Fig.\ref{fig:synthetic}.(a)) is a trimmed version of the target. 
Then we apply to the source a random diffeomorphic deformation $\psi$ that we will try to retrieve by registration.

We have shown in Fig.\ref{fig:synthetic} that the classic distance aims at registering the entire source onto the entire target. 
This leads to abnormal distortions of the source curves that can be observed in the light blue stretched curve in Fig.\ref{fig:synthetic}.(b). 
On the contrary, both partial dissimilarity terms successfully guide the registration of the source onto a subset of the target. 
The main difference between these two terms can be seen in the bifurcations' neighborhoods : the partial matching dissimilarity fails to register the source curves when the normalized partial matching doesn't fall into a flat local minima.
The excess of mass in the target at the bifurcation has no negative effect with the normalization.\\

\noindent\textbf{Registration of a template onto a real vascular tree}. 
Now that we have illustrated the potential of our method on a toy example, we continue on entire vascular trees extracted from real patients datasets.
Finding correspondences between the annotated template and a raw target is of great interest in clinical applications such as interventional radiology, and could provide automatic partial annotation of the imaged vascular tree. 
\begin{figure*}[!t]
\begin{minipage}{.2\linewidth}
\centering
\centering
   \subfloat[\label{fig:template}]{%
      \includegraphics[trim={7cm 4.5cm 5cm 3cm}, clip, width=1\textwidth]{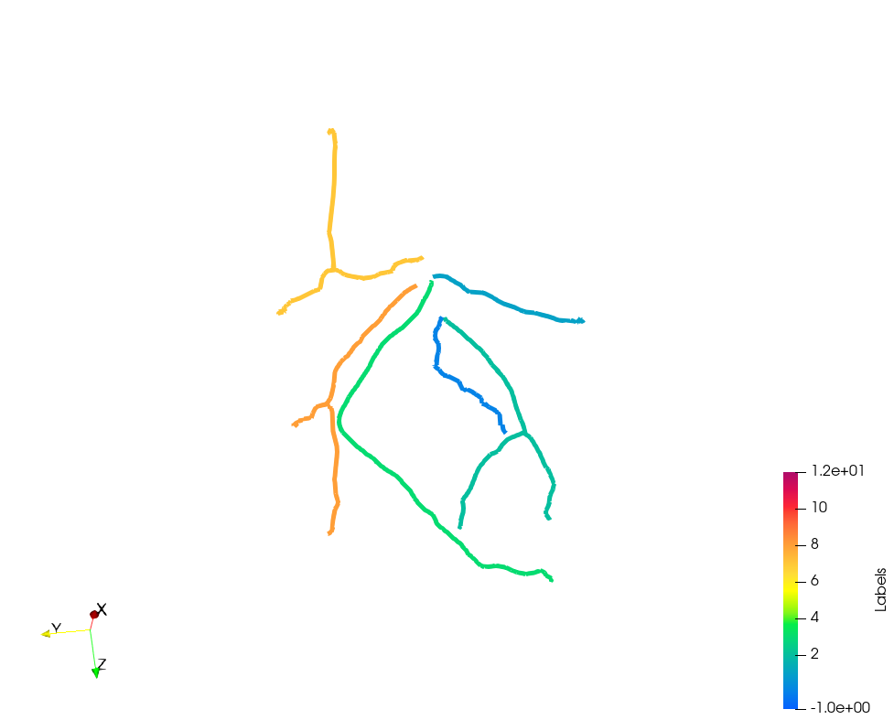}}
\end{minipage}%
\begin{minipage}{.8\linewidth}
\centering
\centering
   \subfloat[\label{fig:Varifold}]{%
      \includegraphics[trim={7cm 3.5cm 5cm 3.5cm}, clip, width=0.28\textwidth]{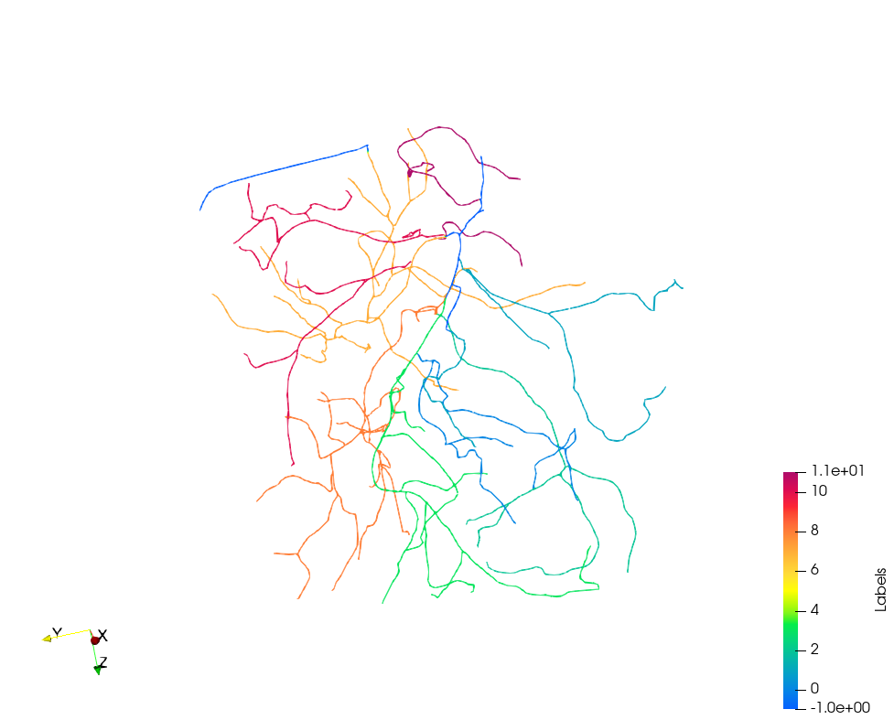}}
\hspace{0.05\textwidth}
\centering
   \subfloat[\label{fig:PartialVarifoldLocal}]{%
      \includegraphics[trim={7cm 3cm 5cm 3.5cm}, clip, width=0.28\textwidth]{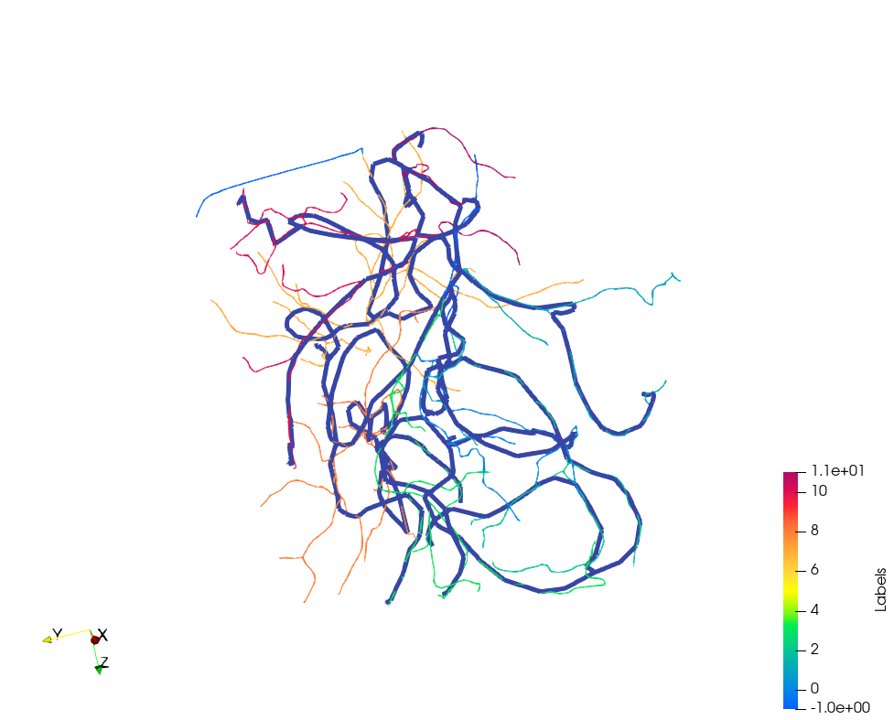}}
\hspace{0.05\textwidth}
\centering
   \subfloat[\label{fig:PartialVarifoldLocalNormalized}]{%
      \includegraphics[trim={7cm 3cm 5cm 3.5cm}, clip, width=0.28\textwidth]{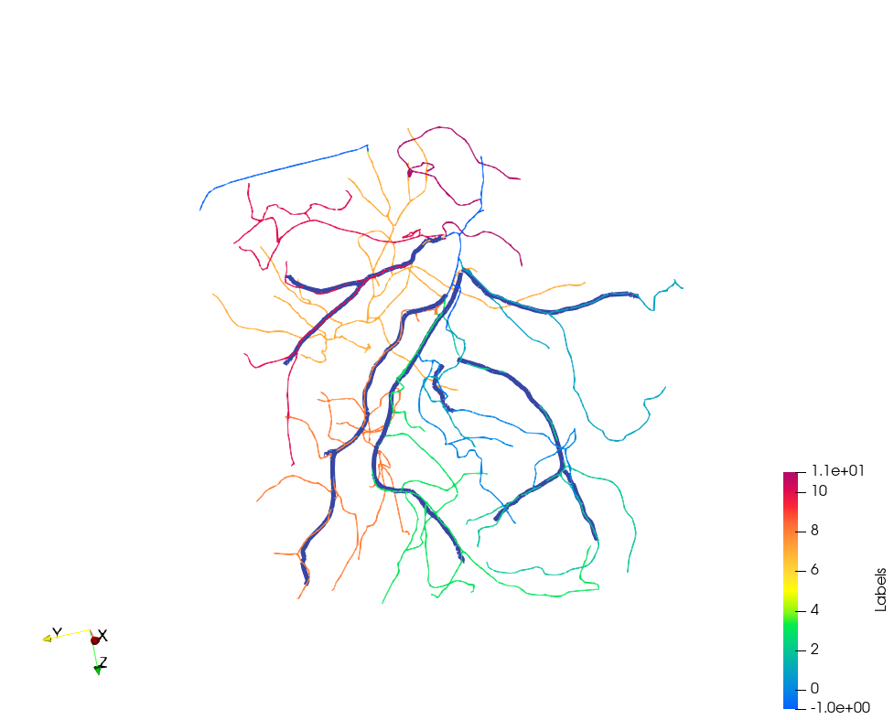}}
\par
\centering
   \subfloat[\label{fig:target2}]{%
      \includegraphics[trim={7cm 2.5cm 5cm 4cm}, clip, width=0.28\textwidth]{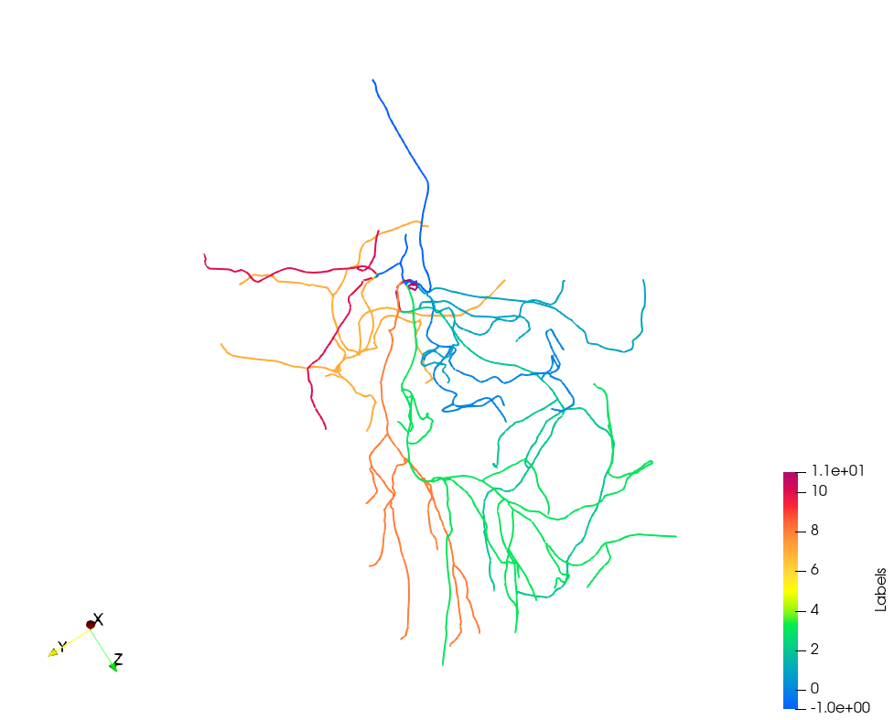}}
\hspace{0.05\textwidth}
\centering
   \subfloat[\label{fig:Varifold2}]{%
      \includegraphics[trim={7cm 2.5cm 5cm 4cm}, clip, width=0.28\textwidth]{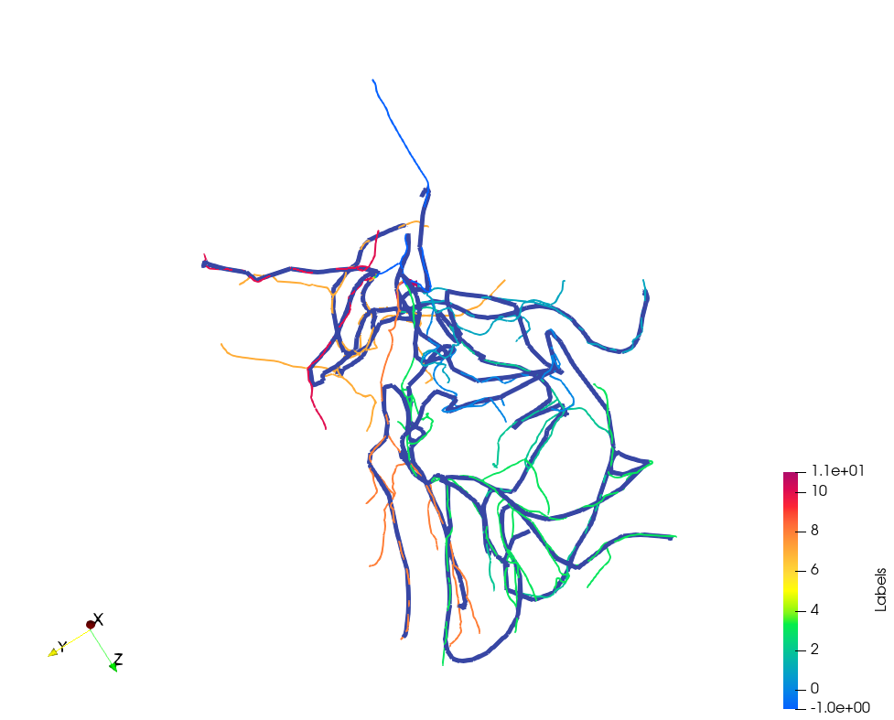}}
\hspace{0.05\textwidth}
\centering
   \subfloat[\label{fig:PartialVarifoldLocalNormalized2}]{%
      \includegraphics[trim={7cm 2.5cm 5cm 4cm}, clip, width=0.3\textwidth]{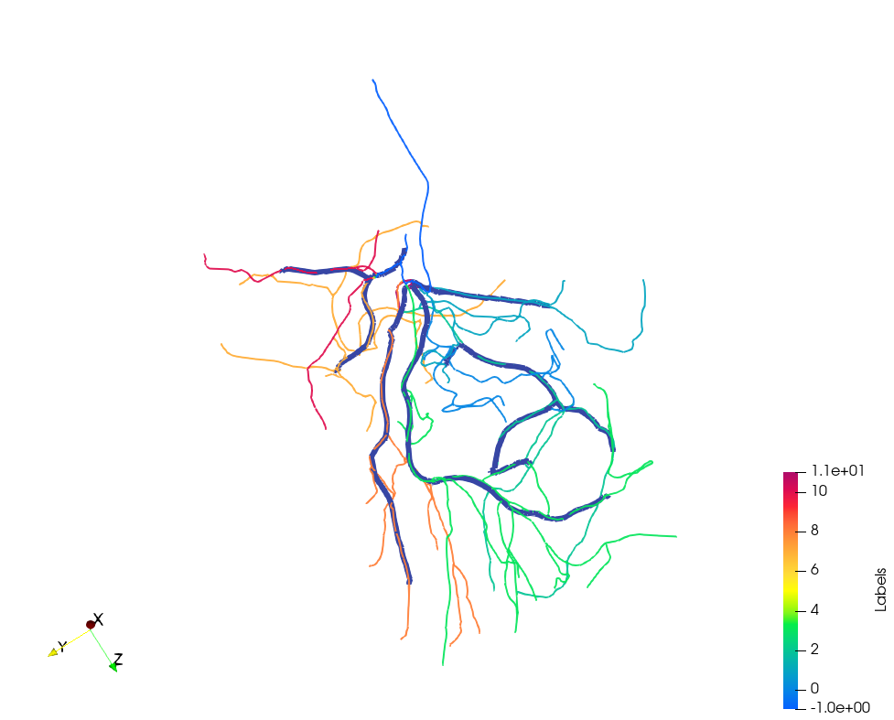}}
\end{minipage}\par\medskip

\caption{\label{fig:tree_comparison} (a) Template composed of different arteries of interest. (b),(e) Real vascular trees. (c),(f) Varifold registrations. (d),(g) Normalized partial matching registrations (Eq. \ref{def:partial_normalized_dissimilarity}). }
\end{figure*}
Because of the high topological variability of the vasculature, there is no perfect template of all the patients, and a simpler template seems suited.
The target tree (Fig.\ref{fig:tree_comparison}. (b), (e)) is obtained by automatic centerlines extraction from an injected CBCT inspired from the fast marching method (\cite{Deschamps2001}) that have been labeled by a clinical expert. 
The source tree (Fig.\ref{fig:tree_comparison}.(a)) is a manually simplified template in which arteries of interest have been selected. 

This experiment is particularly difficult in the classic LDDMM framework, and the partial normalized dissimilarity term introduces a meaningful deformation of the simple template onto a subset of the target despite the wide topological difference.\\

\begin{figure*}[!t]
\centering
   \subfloat[\label{fig:source}]{%
      \includegraphics[trim={4cm 2cm 4cm 2cm}, clip, width=0.4\textwidth]{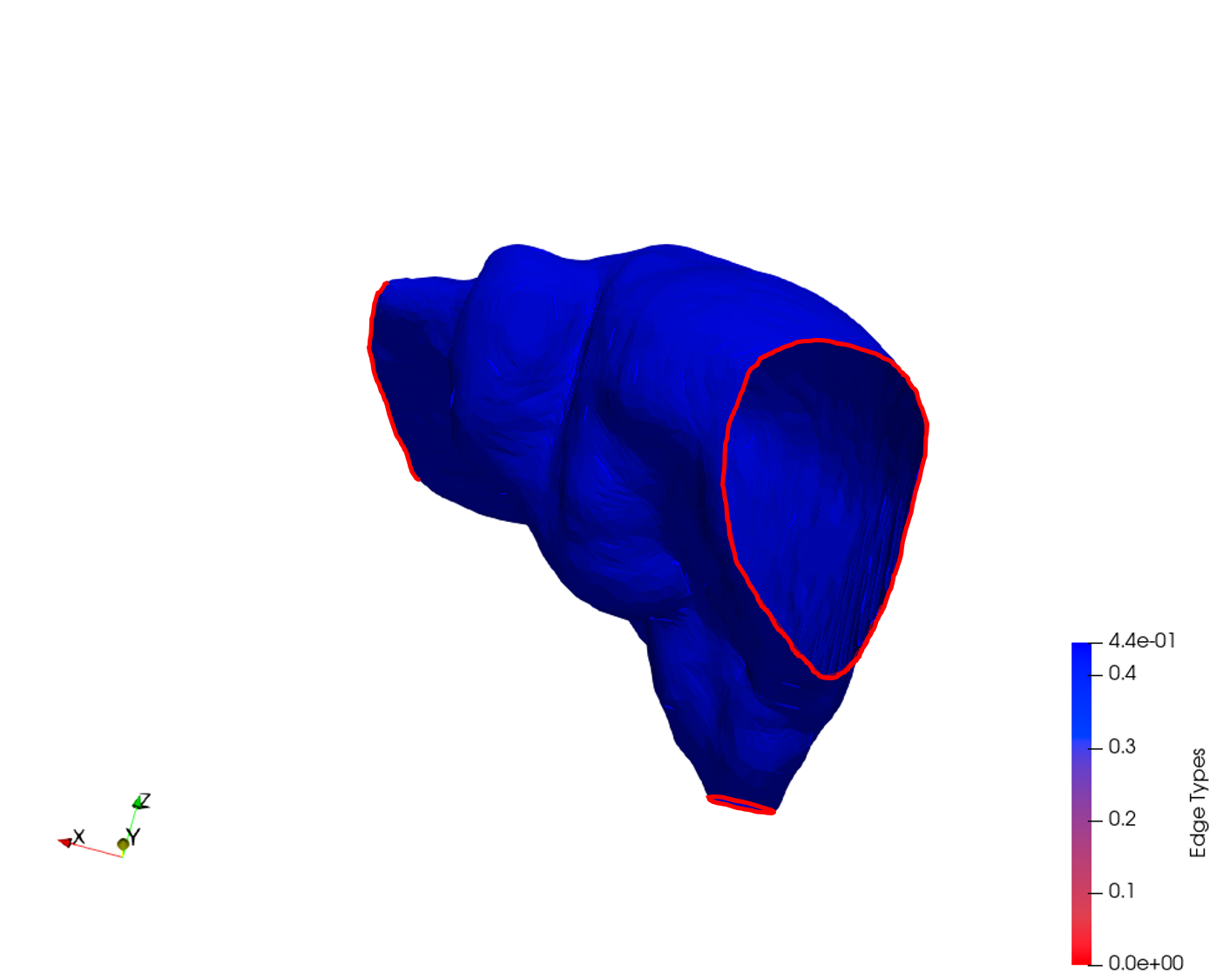}}
\hspace{.5cm}
   \subfloat[\label{fig:target}]{%
      \includegraphics[trim={4cm 2cm 4cm 2cm}, clip, width=0.4\textwidth]{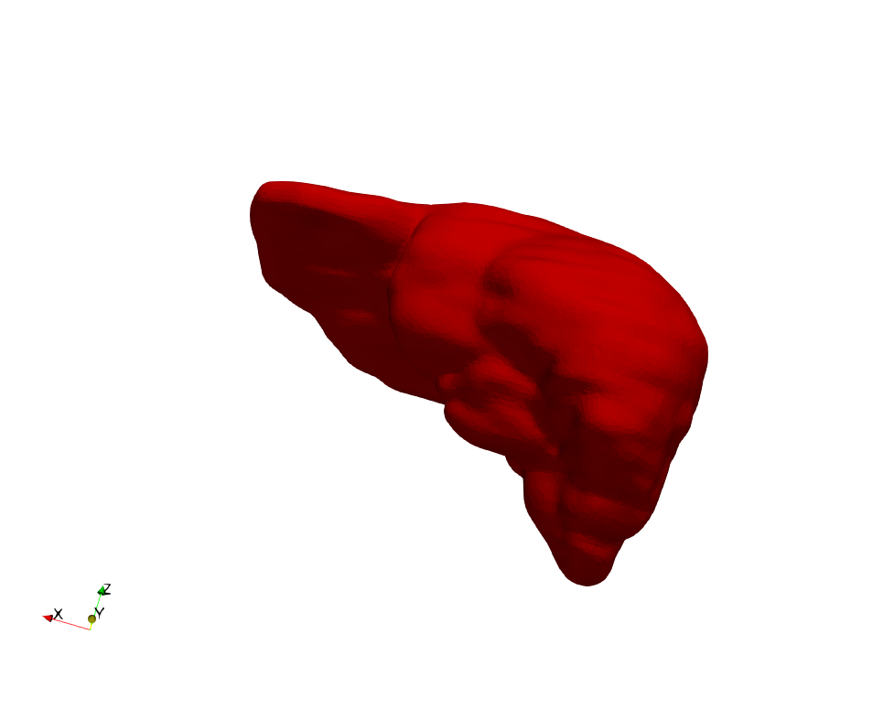}}
\hspace{.5cm}
\centering
   \subfloat[\label{fig:jac_varif}]{%
      \includegraphics[trim={4cm 2cm 4cm 2cm}, clip, width=0.4\textwidth]{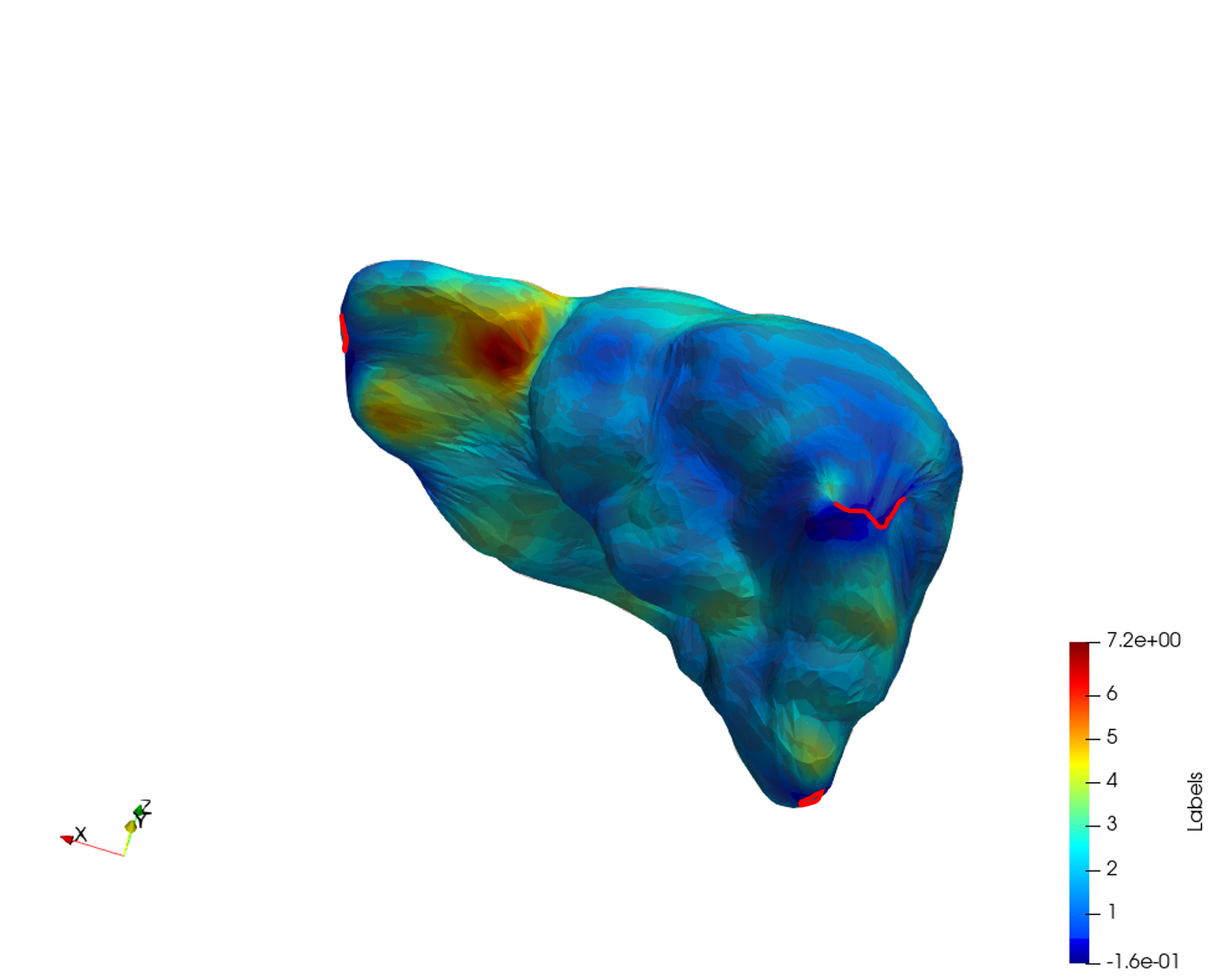}}
\hspace{.5cm}
\centering
   \subfloat[\label{fig:jac_partial_varif}]{%
      \includegraphics[trim={4cm 2cm 4cm 2cm}, clip, width=0.4\textwidth]{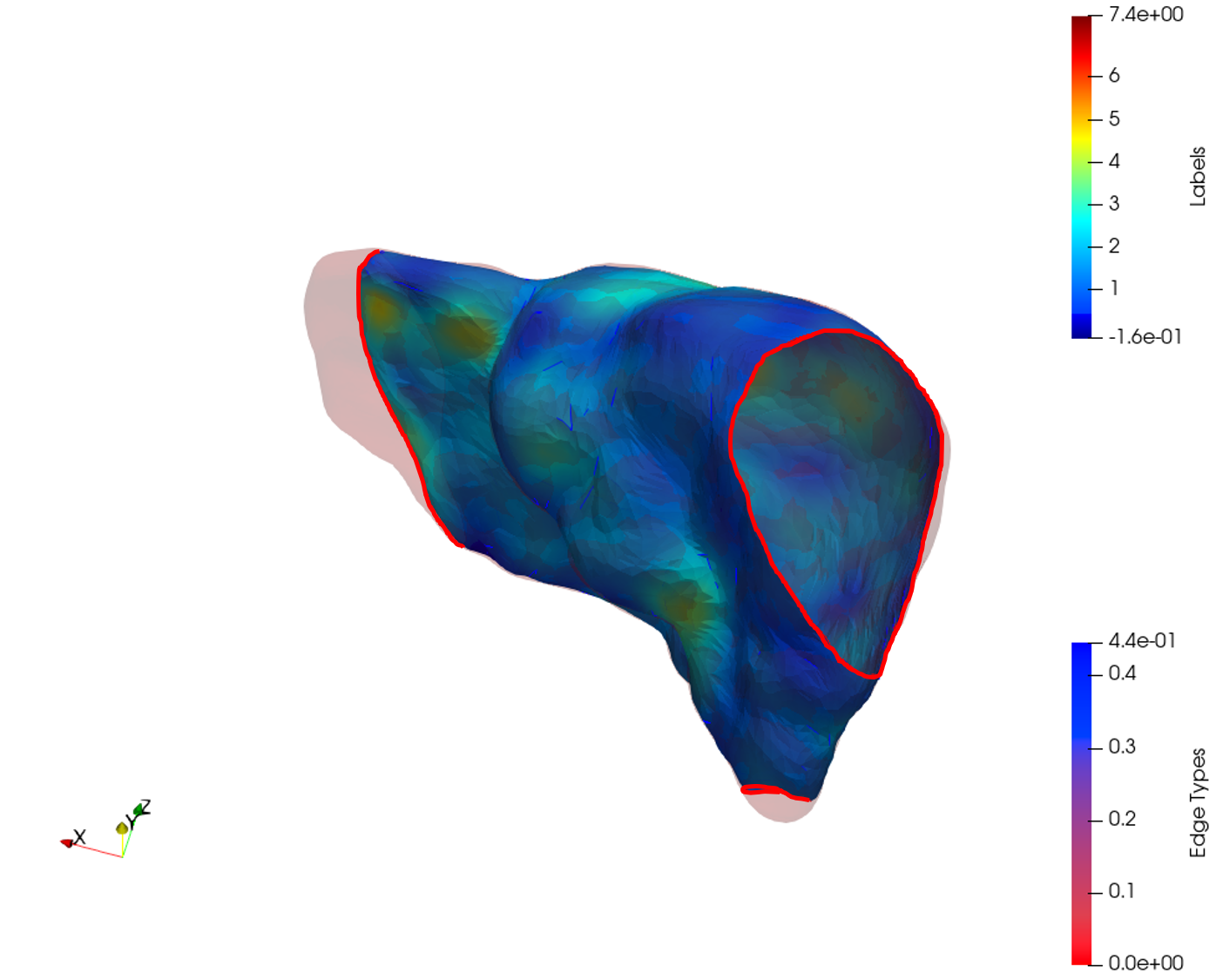}}\\

\caption{\label{fig:Surface_partial_registration_jacob} Registration of a truncated liver's surface from a CBCT (a) onto a complete liver's surface from a CT (b). The deformed surfaces are colored by the determinant of the Jacobian. (c) Varifold registration;  (d) Partial normalized registration (\ref{def:partial_normalized_dissimilarity}) (d).}
\end{figure*}

\noindent\textbf{Liver Surface registration with truncation.}
In the third experiment, we register a truncated liver surface manually segmented from a CBCT onto a complete liver surface manually segmented from a CT scan. 
In CBCT exams, the livers are usually larger than the field of view, causing the truncation of the surface. 
Both acquisitions come from the same patient and are separated in time by one month. 

We show in Fig. \ref{fig:Surface_partial_registration_jacob} the registrations results of the LDDMM associated to the distance in the space of Varifolds and to the normalized partial dissimilarity term. 
The deformed shapes are colored by the determinant of the Jacobian of the deformation, that can be seen as its intensity. 
As expected the Varifold distance leads to unrealistic deformation that tends to fill the holes in the source shape to cover the entire target. 
From the anatomical and medical point of view this is misleading. 
On the contrary the partial matching allows a coherent deformation of the source.

\section{Conclusion}

In this paper we adapted the scalar product of Hilbert spaces of Varifolds to guide partial shape matching and registration. 
 We applied this new data attachment to the registration of a template vascular tree onto real cases.
We also showed that it is suitable for the registration of a truncated surface onto a complete one.
This data term is suited to different shapes comparison such as unions of curves or surfaces.

The main bottleneck of our approach is the risk of shrinkage when no easy registration is possible. A promising lead to tackle this issue would be to also find a subset of the target to include in the source, such as \cite{Bronstein2009}. 
The dissimilarity term introduced can be modified to fit the problem of registering a complete shape onto a truncated one. 
In further work we would also like to extend this dissimilarity term to other space of representations such as the Normal Cycles or the functional Varifolds.

\paragraph{} \textit{Acknowledgement} We would like to thank Perrine Chassat for her early work on the partial matching , allowing us to easily adapt the new data fidelity terms to the real case of liver surfaces registrations. 

\bibliographystyle{splncs04}

\end{document}